\documentclass[letterpaper, 10 pt, journal, twoside]{sty/IEEEtran}

\usepackage[noadjust]{cite} 

\usepackage{mathptmx,amsmath,amssymb,mathtools,xfrac}
\usepackage{scalerel,relsize}
\usepackage{tabstackengine}

\usepackage{apptools}

\usepackage[hang,flushmargin]{footmisc}

\usepackage{graphicx,siunitx,xcolor}
\graphicspath{{../fig/}{../img/}} 
\DeclareGraphicsExtensions{.pdf,.jpeg,.png}

\usepackage{algorithmic,paralist}

 \usepackage{times} 

\usepackage{array,adjustbox,nccmath,afterpage,xspace,multicol}
\usepackage{booktabs} 

\usepackage{etoolbox}
\makeatletter
\patchcmd{\@makecaption}
  {\scshape}
  {}
  {}
\makeatletter
\patchcmd{\@makecaption}
  {\\}
  {.\ }
  {}
\makeatother
\def\tablename{Table}
\usepackage{colortbl}

\usepackage{floatrow}
\usepackage{fixltx2e}
\usepackage{stfloats}

\usepackage{url}

\usepackage{tikz}
\usetikzlibrary{shapes, calc, arrows, shadows, positioning, bending,tikzmark,decorations.pathreplacing}

\usepackage[hidelinks]{hyperref}
\usepackage[capitalize]{cleveref} 
\crefname{equation}{Eqn.}{Eqn.}
\crefname{section}{Sec.}{Sec.}

\newcommand{\minus}{\scalebox{0.65}[1.0]{$-$}}   
\newcommand\myeq{\mkern1.5mu{=}\mkern1.5mu}
\newcommand\mycross{\mkern1.5mu{\times}\mkern1.5mu}

\newcommand\VPa[1][color_blue]{${{VP_{\color{#1}{A}}}}$\xspace}
\newcommand\VPb[1][color_red]{${VP_{\color{#1}{B}}}$\xspace} 
\newcommand\VPbl[1][color_darkred]{${VP_{\color{#1}{BL}}}$\xspace}

\newenvironment {annotatedFigure}[1]{\centering\begin{tikzpicture}[remember picture]
\node[anchor=south west,inner sep=0] (image) at (0,0) {#1};\begin{scope}[x={(image.south east)},y={(image.north west)}]}{\end{scope}\end{tikzpicture}}

\newcommand*\sublabel[4]{
\node at (#2) [fill=none,shape=circle,draw=none, scale=1,inner sep=1pt,font=\sffamily,text=#3] {\text{\relscale{#4}{#1}}};}

\newcommand{\mysup}[1]{\scaleto{\, \mathrm{#1}}{4pt}} 
\newcommand{\mysub}[1]{\scaleto{\, \mathrm{#1}}{4pt}} 
\newcommand{\myisup}[1]{\scaleto{\, \mathrm{#1}}{5pt}} 
\newcommand{\myisub}[1]{\scaleto{\, \mathrm{#1}}{5pt}} 

\definecolor{color_gray}{rgb}{0.5 0.5 0.5}
\definecolor{color_darkgray}{rgb}{0.3 0.3 0.3}
\definecolor{color_red}{rgb}{0.6980    0.0941    0.1686} 
\definecolor{color_darkred}{rgb}{0.3017    0.0004    0.0026} 
\definecolor{color_blue}{rgb}{0.1294    0.4000    0.6745}
\definecolor{color_darkblue}{rgb}{0.0011    0.0472    0.2691} %
\definecolor{color_green}{rgb}{0.3020    0.5725    0.1294}
\definecolor{color_darkgreen}{rgb}{0.0912    0.3278    0.0167}
\definecolor{color_yellow}{rgb}{1.0000    0.8871    0.5453}
\definecolor{color_teal}{rgb}{10.5804    0.8275    0.7765}

\title{\LARGE \bf
A Control Method to Compensate Ground Level Changes \\ for Running Bipedal Robots
}

\author{{\"O}zge Drama and Alexander Badri-Spr{\"o}witz%
\thanks{This work was supported by the International Max Planck Research School for Intelligent Systems (IMPRS-IS).}
\thanks{{\"O}zge Drama and Alexander Badri-Spr{\"o}witz are with the Dynamic Locomotion Group at Max Planck Institute for Intelligent Systems, Stuttgart, Germany {\tt\small \{drama,sprowitz\}@is.mpg.de}}%
}

\begin{document}

\maketitle

\begin{abstract}
%
Bipedal running is a difficult task to realize in robots, since the trunk is underactuated and control is limited by intermittent ground contacts. 
%
Stabilizing the trunk becomes even more challenging if the terrain is uneven and causes perturbations.
%
One bio-inspired method to achieve postural stability is the virtual point (VP) control, which is able to generate natural motion.
 However, so far it has only been studied for level running.
%
%
In this work, we investigate whether the VP control method can accommodate single step-downs and downhill terrains.
We provide guidelines on the model and controller parameterizations for handling varying terrain conditions.
%
Next, we show that the VP method is able to stabilize single step-down perturbations up to \SI{\textbf{40}}{\textbf{\centi\meter}}, and downhill grades up to 20-\SI{\textbf{10}}{\textbf{\degree}}~corresponding to running speeds of 2-\SI{\textbf{5}}{\textbf{\meter\per\second}}.
%
Our results suggest that VP control is a promising candidate for terrain-adaptive running control of bipedal robots.

\end{abstract}

\section{INTRODUCTION} \label{sec:Intro} 

Several bipedal robots are able to walk steadily on flat terrain and use a pose controller that maintains an upright trunk at all times \cite{Ding_2018,Kim_2007,Sheng_2013}.
%
However, sustaining trunk stability becomes difficult under external perturbations such as changes in ground level, since the control mechanism needs to regulate the additional change in system's energy \cite{Tokur_2019}. 
%
Perturbations can be either local, like a single step up/down, or global, as in up/downhill terrain.

Existing solutions to encounter uneven terrain depend on the level of pose and terrain estimation capabilities of the robot.
%
Humanoids that use zero moment point control to walk utilize additional control layers to cope with slopes.
%
A pose controller is used to adjust the ankle pitch to prevent the robot from tilting, especially if the robot has inertial sensor at its torso (e.g.,~Nao \cite{Ding_2018},~KHR-2 \cite{Kim_2007},~SCUT-I \cite{Sheng_2013}).  
The robot SUBO-I has an additional disturbance observer to adjust the robot height w.r.t. the slope \cite{Cho_2018}, and the robot DRB-HUBO has a foot orientation adaptation mechanism for incorporating the effect of the terrain slope \cite{Joe_2019}.
%
The virtual model control mechanism is also able to accommodate slopes by adjusting the desired hip height \cite{Chew_1999}.
%
On the other hand, some robots have vision-based perception, and therefore have extended capabilities to estimate the terrain and react to the changes \cite{Fallon_2015}.
%
%
One common objective of these controllers is that they all maintain a fixed upright trunk throughout the motion.
%
An exception is the SD-2 robot, which moves its trunk to offset the shift of its center of gravity due to the up/downhill slope \cite{Zheng_1990}.
%

Bipedal running has an additional difficulty: large and rapidly changing ground reaction forces destabilize the underactuated trunk and the controller has less time to regulate the system during stance \cite{Nilsson_1989,Gottschall_2005}.
The essential properties of bipedal running are captured by the spring loaded-inverted pendulum model with a trunk  (TSLIP).
Within the TSLIP framework, virtual point (VP) control is proposed as a mechanism to achieve postural stability \cite{Maus_2008}, which is implemented in the ATRIAS robot for walking \cite{Peekema_2015}.
The VP approach forms a geometric coupling between the leg force and hip torque, based on the assumption that the ground reaction forces (GRF) intersect at a point above, at, or below the center of mass (CoM). 
%
%
The method is explored extensively for level running \cite{Andrada_2014, Drama_2019,Drama_2020, Sharbafi_2012}. 
%
However, there is no formalism to describe how VP control can be used to accommodate varying terrain conditions.
So far, a single study suggests to horizontally offset the VP position proportional to the change in step size to traverse stairs and slopes as a concept~\cite{Kenwright_2011}.

In this paper, we aim to explore model and controller parameterizations within the TSLIP-VP control framework to accommodate varying terrain conditions.
In the first part of our work, we investigate whether the VP control mechanism can counteract external perturbations introduced by a single drop in the ground level. 
%
%
 In the second part, we search for feasible ways to use the VP and achieve stable locomotion patterns for downhill running. 
The decrease in ground level adds energy to the system, equal to the change in potential energy. For the biped to maintain constant speed, it is necessary to adjust the posture and leg parameters (i.e., leg length, leg stiffness and damping, leg damping, leg angle at touch-down).
We formalize which adjustments are sufficient for the TSLIP and the VP control scheme.
The resulting insights can be used to efficiently parameterize control mechanisms that allow bipedal robots to compensate for ground level changes.

\subsection{Related Work in Biomechanics} \label{sec:RelWork} 

In order to extend the VP concept in a feasible and efficient manner, we take insights from human locomotion and analyze how humans cope with terrain changes.
Humans adjust their leg properties and posture to respond to the changes in the ground level during running. 
In the presence of a visible single drop in ground level, humans adjust their leg parameters during the prior and at the perturbed step \cite{Mueller_2012}. 
They decrease their leg stiffness, increase their leg angle, and elongate their leg at touch-down \cite{Mueller_2010,Mueller_2012}. 
The peak GRF decreases at the preparation step, followed by an increased peak GRF at the perturbed step \cite{Mueller_2012}.
The GRF vectors intersect at a virtual point below the center of mass (\VPb), whose magnitude is reported as \SI{30}{\centi\meter} for running over a ground level drop of \SI{10}{\centi\meter} at \SI{5}{\meter\per\second} \cite{Vielemeyer_Drama_2020}.
If the perturbation is visually hidden from the subjects (i.e., camouflaged), the adaptations in leg parameters are similar to those of the visible setting in principle, but display a larger behavioral variance between subjects. The vertical location of the estimated VP shows a larger variation for the camouflaged drop as well \cite{Vielemeyer_Drama_2020}.

Terrain with a downhill slope can be modeled as a combination of subsequent ground level drops.
%
The biomechanical literature for downhill running involves slopes up to -\SI{20}{\percent} and running speeds up to \SI{5}{\meter\per\second} \cite{Vernillo_2017}.
%
%
In terms of temporal gait parameters, downhill running yields an increased aerial time, reduced step frequency and decreased duty factor compared to level running \cite{Vernillo_2017}.
Human runners also adjust their postural orientation at heel strike to accommodate downhill terrain. 
The authors of \cite{Chu_2004} report two separate postural responses, where the first group of participants showed a more extended posture with low shock attenuation and the second participant group showed a more flexed posture with high shock attenuation.

Observation of the GRF patterns and the body's center of mass (CoM) energetics provides insights about the kinetic adaptations humans utilize for downhill running.
%
%
The impact peak of the vertical ground reaction forces increases with the downhill slope, whereas the active peak either remains identical \cite{Dick_1987,Gottschall_2005,Telhan_2010} or decreases \cite{Wells_2018}.
In addition, the maximum vertical GRF shifts from the active to the impact peak, as downhill slope increases \cite{Wells_2018}.
%
There are two different trends that are reported for the peak horizontal GRF during downhill running, which we summarize in  \cref{tab:literature}. 
The authors of \cite{Dick_1987,Gottschall_2005,Wells_2018} report an asymmetric gait behavior, where the peak propulsion forces become higher and peak braking forces become lower. 
Other studies \cite{Telhan_2010,Yokozawa_2005} suggest that peak horizontal GRF remain the same.
In downhill running, the external mechanical work (i.e. the work done to move the body’s CoM with respect to the environment) is reported to be positive (i.e. energy generation) at shallow grades below -\SI{10.5}{\percent} and negative (i.e. energy dissipation) at steeper grades~\cite{Snyder_2011}.
%

\begin{table}[tb!]                                                                                                                                                                                                                                                                                                                                                                                                                                                                                                                                                                 
\centering                                                                                                                                                                                                                                                                                                                                                                                                                                                                                                                                                                          
\captionsetup{justification=centering}
\caption{Downhill running experiments reported in the literature. Percentage values with a plus sign indicate an increase and a minus sign indicates a decrease in magnitude compared to the level running conditions. The slope percentage is calculated as the rise of the slope divided by its run times 100.}
\label{tab:literature}
\begin{adjustbox}{width=0.98\textwidth}
\begin{tabular}{@{} l  c c c c c c @{}}
\multicolumn{1}{l}{\hphantom{u}} & \multicolumn{1}{c}{} & \multicolumn{2}{c}{Peak vert. GRF} &   \multicolumn{2}{c}{Peak horz. GRF} &\multicolumn{1}{c}{} \\
\cline{3-4} \cline{5-6}
\multicolumn{1}{l}{\hphantom{u}Speed} & \multicolumn{1}{c}{Slope Grade} & Impact & Active &  Braking & Propulsion &\multicolumn{1}{c}{Reference} \\
\hline 
\hspace{1mm} \SI{4.5}{\meter\per\second}  & -\SI{8.5}{\percent} & +14\,\si{\percent}   & no change & double the &  half the & \cite{Dick_1987}\\
\hspace{1mm}  &  &   & \multicolumn{2}{r}{prop. force} &   \multicolumn{2}{l}{braking force} \\
\arrayrulecolor{gray}\hline  
\hspace{1mm}   & -\SI{3}{\degree} & +\SI{18}{\percent} &  & +\SI{27}{\percent}  & -\SI{22}{\percent}  & \\
\hspace{1mm}  \SI{3}{\meter\per\second} & -\SI{6}{\degree} & +\SI{32}{\percent}  &  no change & +\SI{46}{\percent} & -\SI{40}{\percent}  & \cite{Gottschall_2005} \\
\hspace{1mm}  & -\SI{9}{\degree} & +\SI{54}{\percent} &  & +\SI{73}{\percent}  & -\SI{61}{\percent}  & \\
\arrayrulecolor{gray}\hline  
\hspace{1mm} 3,4,5 \si{\meter\per\second} & -3,\,-6,\,-9\,\si{\percent} & higher & -- & \multicolumn{2}{c}{no change} & \cite{Yokozawa_2005}   \\
\arrayrulecolor{gray}\hline 
\hspace{1mm} \SI{3}{\meter\per\second} & -\SI{4}{\degree} & higher & no change &  \multicolumn{2}{c}{no change} & \cite{Telhan_2010} \\
\arrayrulecolor{gray}\hline 
\hspace{1mm} \SI{2.7}{\meter\per\second} & -2,\,-5,\,-8\,\si{\percent} & \multicolumn{2}{c}{higher max vert. GRF} &  \multicolumn{2}{c}{--} & \cite{Lussiana_2015} \\
\arrayrulecolor{gray}\hline 
\hspace{1mm}  & -\SI{5}{\percent} &  +\SI{14}{\percent} &  -\SI{1}{\percent} &  +\SI{2}{\percent} & +\SI{3}{\percent}  & \\
\hspace{1mm} \SI{4}{\meter\per\second} & -\SI{10}{\percent} & +\SI{32}{\percent}   &  -\SI{3}{\percent} & +\SI{2}{\percent} & no change & \cite{Wells_2018} \\
\hspace{1mm}& -\SI{15}{\percent} &  +\SI{47}{\percent}  &  -\SI{6}{\percent} & +\SI{10}{\percent} & -\SI{5}{\percent}  & \\
\hspace{1mm}  & -\SI{20}{\percent} & +\SI{61}{\percent}   &  -\SI{8}{\percent} &  +\SI{5}{\percent} & -\SI{13}{\percent}  & \\
\end{tabular}
\end{adjustbox}
\end{table}                                                                                                                                                                                                                                                                                                                                                                                                                                                                                                                                                                         
\raggedbottom

\section{SIMULATION MODEL} \label{sec:SimModel}

\begin{figure}[!t]
\centering
\begin{annotatedFigure}
	{\includegraphics[width=0.98\linewidth]{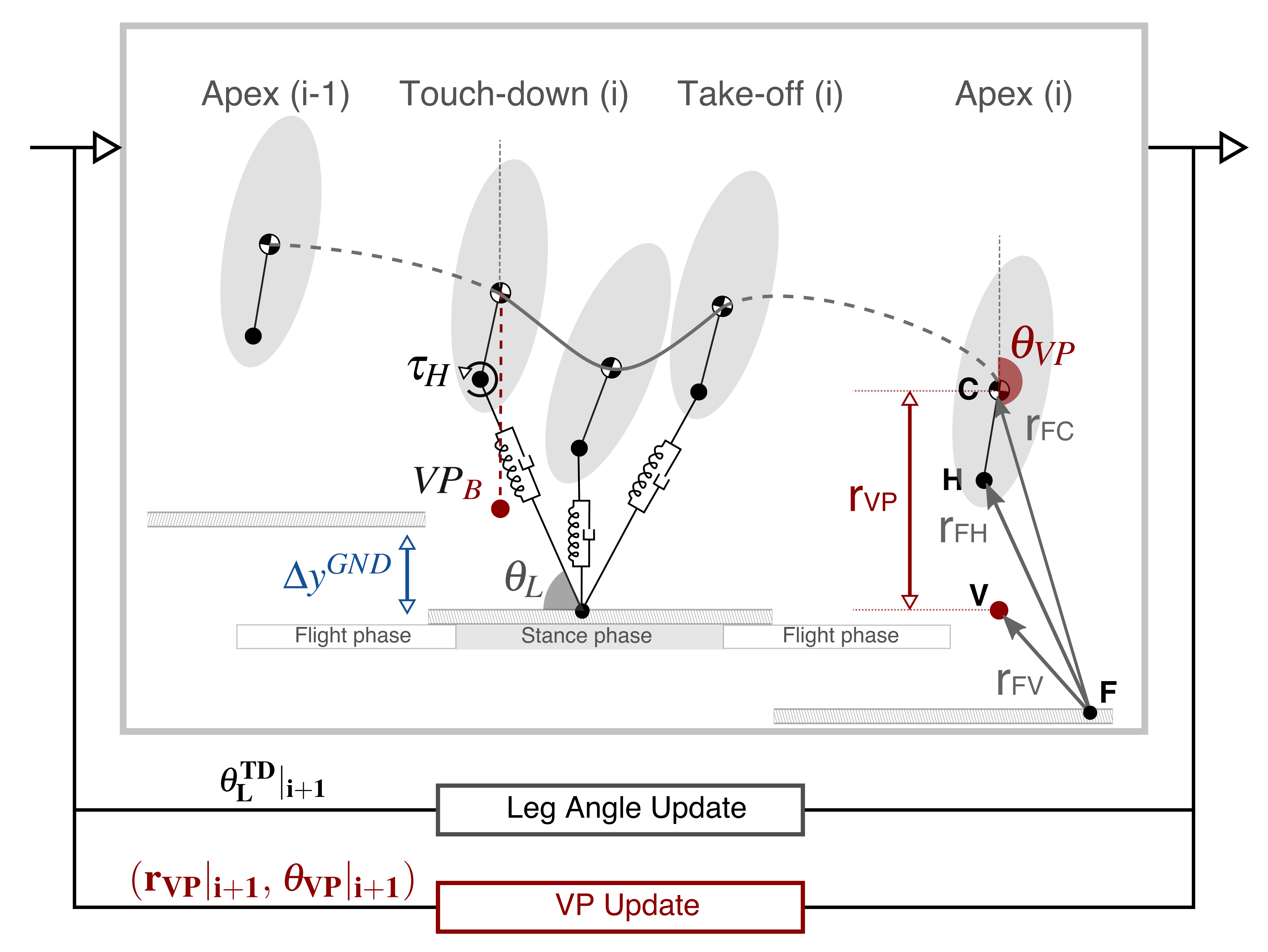}}
\end{annotatedFigure}
\caption{The TSLIP model has hybrid dynamics, which involve a flight phase followed by a stance phase. During the stance phase, the dynamics of the leg is passive, whereas the hip is actuated with a torque. The hip torque is defined in a way that the resultant ground reaction forces point to a virtual point. It is located \SI{30}{\centi\meter} below the center of mass (\VPb) in our setup. The leg angle $\theta_{L}$ and VP angle $\theta_{VP}$ are updated at the end of each step. At each step, the ground level drops by $\Delta y^{\mysup{GND}}$ to simulate downhill running.
%
}\label{fig:TSLIP}
\end{figure}

In this section, we describe the TSLIP model that we use to investigate the VP as a control scheme for accommodating ground level changes.
The TSLIP model consists of a trunk with mass $m$ and moment of inertia $J$, which is attached to a massless leg of length $l$, as shown in \cref{fig:TSLIP}.
The values for these model parameters are taken from \cite{Drama_2019}.

The leg consists of a parallel spring-bilinear damper mechanism, where the sum of spring ($F_{\mathrm{sp}}$) and damping ($F_{\mathrm{dp}}$) forces equal to the axial component of the GRF ($ \prescript{}{F}{\mathbf{F}}_{a}$).
The hip is actuated with a torque ($\tau_{H}$),  which generates the tangential component of the GRF ($\prescript{}{F}{\mathbf{F}}_{t}$) expressed as,
%
\begin{equation}
\begin{aligned}
&\prescript{}{F}{\mathbf{F}}_{a} =\overbrace{( \, k \, (l \minus l_{0})}^{\scaleto{F_{\mathrm{sp}}}{10pt}} \, \minus \, \overbrace{c \, \dot{l} \, (l  \minus l_{0}) \,)}^{\scaleto{F_{\mathrm{dp}}}{10pt}}   \times  \bracketVectorstack{\minus \cos\theta_{L} \\ \hphantom{s} \sin\theta_{L}} \\
&\prescript{}{F}{\mathbf{F}}_{t} = \minus l^{\minus1} \times \underbrace{ \prescript{}{F}{\mathbf{F}}_{a} \times  \left[   \frac{\mathbf{r}_{FV} \times \mathbf{r}_{FH} }{\mathbf{r}_{FV} \cdot \mathbf{r}_{FH}}\right]   \times  l}_{\scaleto{\tau_{H}}{7pt}} \times \bracketVectorstack{\hphantom{s}\sin\theta_{L} \\  \minus \cos\theta_{L} }.  \\ 
\end{aligned}
\label{eqn:tauVP}
\end{equation}
The leg spring-damper jointly dissipate energy from the system, whereas the hip actuator supplies an equal amount of energy to preserve the energy balance.
The hip torque is defined through a virtual point with radius ($r_{VP}$) and angle ($\theta_{VP}$). Placing the VP above (\VPa) or below (\VPb) the center of mass affects the pattern of trunk angular motion\footnote{The virtual points that are below the CoM and below the leg axis are referred to as \VPbl in \cite{Drama_2019,Drama_2020}. In this paper, the \VPb radius is -\SI{30}{\centi\meter} and is always below the leg axis.}. The \VPa is defined with respect to the \emph{body frame}, which is centered at the CoM and is aligned with the trunk. On the other hand, \VPb is defined with respect to the \emph{world frame}, which is centered at the CoM and is aligned with the global vertical axis\footnote{\;The body frame translates with the CoM and rotates with the trunk. The world frame translates with the CoM and does not rotate.} \cite{Drama_2020}.  
 
%
The simulation starts at the apex state with zero vertical acceleration, which is followed by a flight phase with ballistic dynamics. The stance phase begins with the leg touch-down, during which the equations of motion for the CoM state $(x_{C},  y_{C},  \theta_{C})$ are expressed as,
\begin{equation}
m \begin{bmatrix}  \ddot{x}_{C} \\  \ddot{z}_{C} \end{bmatrix}  \myeq \prescript{}{F}{\mathbf{F}}_{a} +  \prescript{}{F}{\mathbf{F}}_{t} + g, 
\hspace{0.1cm} \text {and} \hspace{0.1cm}
J\, \ddot{\theta}_{C}  \myeq \minus {\mathbf{r}}_{FC} \mycross (  \prescript{}{F}{\mathbf{F}}_{a} +  \prescript{}{F}{\mathbf{F}}_{t}).
\label{eqn:EoM}
\end{equation}
The stance phase ends when the leg reaches to its rest length $l_{0}$, the vertical GRF becomes zero, or the vertical CoM acceleration becomes zero after the mid-stance.

%

\section{PROPOSED CONTROL METHOD} \label{sec:PropMethod}
The leg angle at touch-down $\theta_{\mysub{L}}^{\mysup{TD}}$, VP radius $r_{VP}$ and angle $\theta_{VP}$ are linearly adjusted at the apex of each step, as in \cref{fig:TSLIP}. 

 \subsection{Leg Angle Control}\label{subsec:cont_th0}
The main purpose of the leg angle control is to achieve the desired forward speed and to assist maintaining the desired trunk pitch angle. 
The VP angle control influences the system's energy regulation indirectly by adjusting the coupling between the leg and hip. The hopping height is not directly controlled, but the difference in subsequent apex heights is a control term in the leg angle control \cref{eqn:thetaL0} to assist stability. 

We adjust the desired leg angle at touch-down at each apex of step step $i$ based on the linear control scheme,
\begin{equation}
\begin{aligned}
 \theta_{\mysub{L}}^{\mysup{TD}} \, |_{\myisub{i}} & =  \theta_{\mysub{L}}^{\mysup{TD}} \, |_{\myisub{i \minus 1}}  
  + k_{y} (\Delta {y}_{\mysub{C}}^{\mysup{\, AP}} \, |_{\myisub{i \minus 1}}^{\myisup{i}} + \Delta{y_{\mysub{GND}}}|^{\myisup{i}}_{\myisub{i\minus1}})
 \\
 & + k_{\dot{x}_{0}}  ( \dot{x}_{\mysub{C}}^{\mysup{DES}} - \dot{x}_{\mysub{C}}^{\mysup{\, AP}} \, |_{\myisup{i}} )
    +  k_{\dot{x}}  (\Delta \dot{x}_{\mysub{C}}^{\mysup{\, AP}} \, |_{\myisub{i  \minus 1}}^{\myisup{i}} ) \\
 & +  k_{\theta} \,||\, {\theta}_{\mysub{C}}^{\mysup{DES}} - {\theta}_{\mysub{C}}^{\mysup{AP}} \, |_{\myisub{i}} \,||\,
 +  k_{\bar{\theta}} ( {\theta}_{\mysub{C}}^{\mysup{DES}} - { \bar{\theta}}_{\mysub{C}}^{\mysup{AP}} \, |_{\myisub{i\minus1}}^{\myisub{i}}), 
 \end{aligned}
  \label{eqn:thetaL0} 
\end{equation} 
where $(k_{\dot{x}},k_{\dot{x}_{0}})$ components regulate the forward speed, $(k_{\theta},k_{\bar{\theta}})$ components bound the oscillations of the trunk, and $k_{y}$ component guides the stabilization in height. In our notation, $\Delta$ denotes the difference and superscript bar denotes the average value. 
%
If the terrain involves a downhill slope, we include the deviation of the mean trunk angle during stance from the desired trunk angle.
In the course of adjusting the gains of the leg angle controller, we make sure to achieve the desired forward speed and mean trunk angle in a smooth fashion, while excluding non-periodic and period-n trajectories.

 \subsection{Virtual Point Radius and Angle Control}\label{subsec:cont_thVP}
We adjust the VP radius as a function of the angular velocity at leg take-off $\Delta\dot{\theta}_{\mysub{C}} \, |_{\mysup{t=0}}^{\mysup{TO}}$\,, and the VP angle based on the difference between the desired mean body angle $ \theta_{C}^{Des}$ and mean body angle observed in the last step $\Delta\theta_{C}$ as,
 
\begin{subequations}
\begin{align}
\label{eqn:rVP_LIN} 
r_{\mysub{VP}} \, |_{\myisub{i}} & \myeq 
\begin{cases}
    r_{\mysub{VP}} \, |_{\myisub{i-1}} + r^{\prime}_{\mysub{VP}}&  \hspace{-0.1cm} \text{if }  {i} \myeq {i_{\mysub{SD}}}\\
   \frac{r_{\mysub{VP}} \, |_{\myisub{i-1}} + \max \left( 0,\, r_{\mysub{VP}}^{\mysup{DES}} \minus \,\,  k_{r_{\mysub{VP}}} \,\,  || \, \Delta\dot{\theta}_{\mysub{C}} \, |_{\mysup{t=0}}^{\mysup{TO}} || \,  \right )}{2}  & \text{otherwise} 
\end{cases}
\\ 
\label{eqn:thVP_LIN} 
\theta_{\mysub{VP}} \, |_{\myisub{i}} &  \myeq 
\begin{cases}
    \theta_{\mysub{VP}} \, |_{\myisub{i-1}} + \theta^{\prime}_{\mysub{VP}} & \hspace{-0.05cm}  \phantom{templ} \text{if } {i} \myeq {i_{\mysub{SD}}}\\
 \theta_{\mysub{VP}}^{\mysup{DES}}  + k_{\theta_{\mysub{VP}}} \left(  \theta_{\mysub{C}}^{\mysup{DES}}  \minus \,\, \Delta\theta_{\mysub{C}} \, |_{\mysup{TD}}^{\mysup{TO}}   \right )  & \phantom{temp} \text{otherwise}. \\ 
\end{cases}
\end{align}
\label{eqn:rVPthVP_LIN} 
\end{subequations}

The VP adjustment takes place at the end of the step, at apex. If there is a change in the ground level, the VP controller reacts to the changes with one step delay. 
This delayed response poses no problem for downhill running, since the model and control parameters are already tuned to compensate a step-wise continuous perturbation introduced by the global down-slope. 
However, this is not the case for running over a terrain with a single step-down, where the control parameters are adjusted for flat terrain conditions.
The sudden external perturbation might deviate the state excessively, if there is no appropriate response during stepping down.
In particular at slow speeds, the trunk flexion/extension during the step-down (step ${i_{\mysub{SD}}}$) might become too large with the increase in the stance time, and the controller might not recover the state in the following steps. 
To address this issue and reduce the angular rotation during step-down, we propose to offset the VP reference by $(r^{\prime}_{\mysub{VP}}, \theta^{ \prime}_{\mysub{VP}})$ at the end of step $i\minus1$ in \cref{eqn:rVPthVP_LIN}.
%
%

\subsection{Gait Generation and Simulation Configuration}\label{subsec:GaitGen}
\begin{figure}[!t]
\centering
\begin{annotatedFigure}
	{\includegraphics[width=0.98\linewidth]{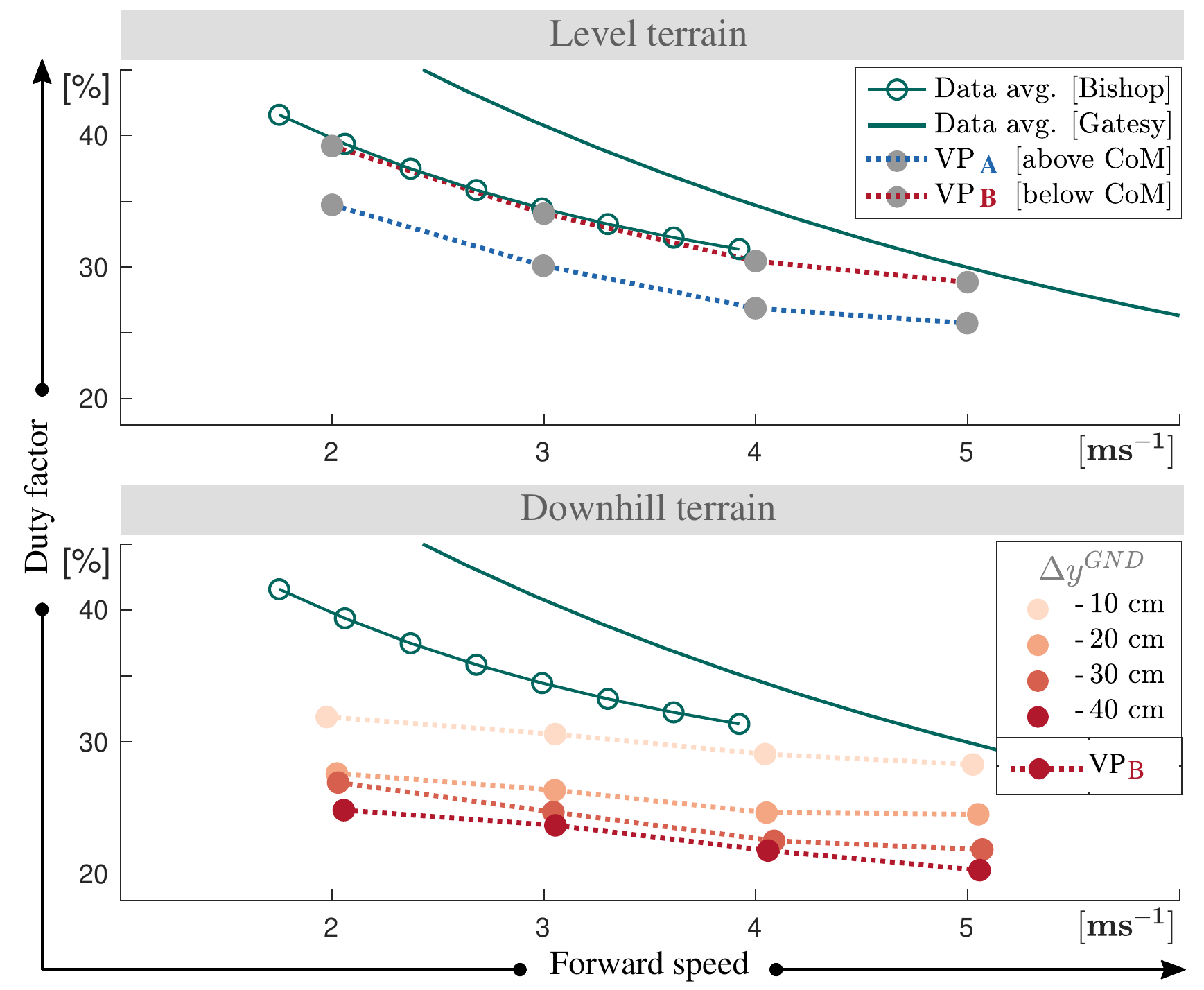}}
	\sublabel{a)}{0.08,0.96}{color_gray}{0.6}    \sublabel{b)}{0.08,0.48}{color_gray}{0.6} 
\end{annotatedFigure}
\caption{Duty factor values of the gaits for the level (a) and downhill (b) terrain conditions.
If the terrain involves a single step-down, the gaits yield duty factor values equal to the duty factor values at level terrain.
We tune the controller parameters and the damping coefficient, in a way that the resultant gaits yield duty factors similar to ones observed in human running.
We also preserve the functional relation that the duty factor decreases with increasing running speed.   
For downhill running the duty factor values get lower and the decrease is proportional to the downhill grade.
%
}\label{fig:duty}
\end{figure}

%
Our simulation study explores two different terrain conditions. 
In the first set of experiments, the terrain involves a single step-down perturbation. 
We conduct a parameter sweep spanning step-down heights of $\Delta y^{\mysup{STP}}\myeq\,$[-10,\,-20,\,-30,\,-40]\,\si{\centi\meter} and speeds of $\dot{x}_{c} \myeq$[2,\,3,\,4,\,5]\,\si{\meter\per\second}. 
We perform the sweep for both VP above (\VPa) and below (\VPb) the CoM, where we set the VP radius to \SI{30}{\centi\meter} based on \cite{Vielemeyer_Drama_2020}.
%

\begin{table}[b!]                                                                                                                                                                                                                                                                                                                                                                                                                                                                                                                                                                 
\centering                                                                                                                                                                                                                                                                                                                                                                                                                                                                                                                                                                          
\captionsetup{justification=centering}
\caption{Terrain slope corresponding to the ground level change per step $\Delta y^{\mysup{GND}}$  and running speed $\dot{x}_{C}$ for \VPb gaits.}
\label{tab:slopeperspeed}
\begin{adjustbox}{width=0.65\textwidth}
\begin{tabular}{@{} l | c c c c  @{}}
\multicolumn{1}{l}{Running } & \multicolumn{4}{c}{Ground level change per step ($\Delta y^{\mysup{GND}}$)}   \\
\cline{2-5}
\multicolumn{1}{l}{\hphantom{u}Speed} & \multicolumn{1}{c}{-\,10 cm} &  \multicolumn{1}{c}{-\,20 cm} &   \multicolumn{1}{c}{-\,30 cm} & \multicolumn{1}{c}{-\,40 cm}  \\
\hline 
\hspace{1mm} 2 \si{\meter\per\second} & 7.2\,\si{\degree} & 12.1\,\si{\degree} & 16.8\,\si{\degree} & 21.5\,\si{\degree} \\
\hspace{1mm} 3 \si{\meter\per\second} & 5.3\,\si{\degree} & 9.0\,\si{\degree} & 12.4\,\si{\degree} & 15.4\,\si{\degree}  \\ 
\hspace{1mm} 4 \si{\meter\per\second} & 4.4\,\si{\degree} & 7.3\,\si{\degree} & 9.7\,\si{\degree} & 12.4\,\si{\degree}   \\
\hspace{1mm} 5 \si{\meter\per\second} & 4.0\,\si{\degree} & 6.6\,\si{\degree} & 8.6\,\si{\degree} & 10.5\,\si{\degree} \\
\end{tabular}
\end{adjustbox}
\vspace{-2mm}
\end{table}

\begin{figure*}[!t]
\centering
\begin{annotatedFigure}
	{\includegraphics[width=0.98\linewidth]{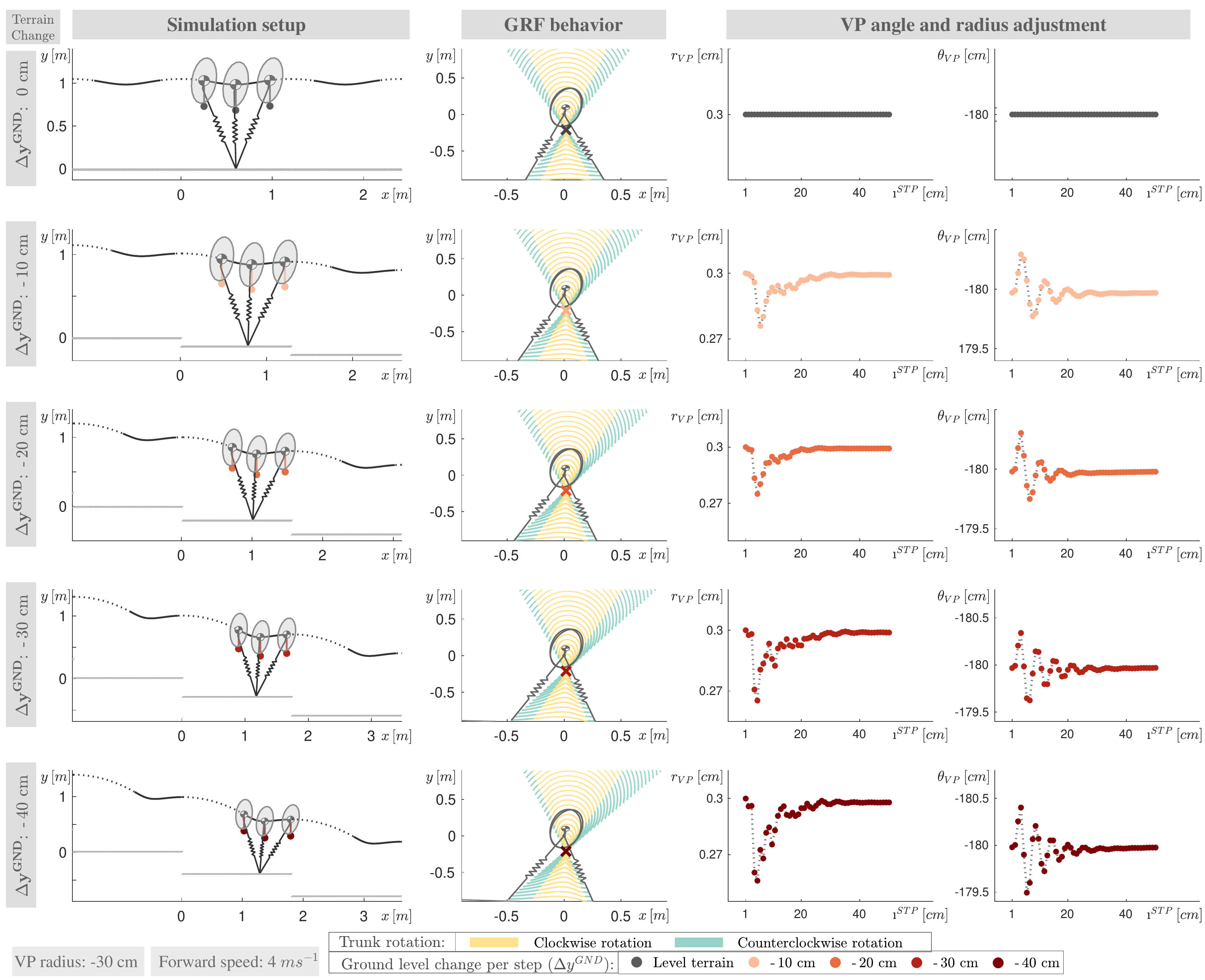}}
	\sublabel{a)}{0.07,0.975}{color_gray}{0.6} 
	 \sublabel{b)}{0.395,0.975}{color_gray}{0.6} 
	  \sublabel{c)}{0.62,0.975}{color_gray}{0.6} 
\end{annotatedFigure}
\caption{a) The simulation setup for ground level change ranging from \SI{0}{\centi\meter} to -\SI{40}{\centi\meter} ground level drop per step ($\Delta y_{\mysup{GND}}$). b) The ground reaction force lines of the converged gaits (50\textsuperscript{th}~step) are shown with dotted lines, where the GRF lines corresponding a clockwise trunk rotation are shown in yellow color and, in teal color otherwise. The VP is marked with a cross. The distribution of the teal-yellow colored areas changes with the terrain grade, which corresponds the change in the trunk motion pattern. c) The VP radius and angle modulation, as the gait reaches to its steady state condition. 
%
}\label{fig:GaitGen}
\end{figure*}

In the second set of experiments, we simulate a downhill slope by deceasing the ground level by a constant amount of ($\Delta y^{\mysup{GND}}$) at the apex of each step.
The slope of the terrain depends on the running speed, which is provided in \cref{tab:slopeperspeed}.
%
%
For downhill running, we focus on using \VPb as the control target, as it is the behavior that is observed in human running. While it may be possible to adjust additional model parameters and the control strategy to use \VPa, we found the \VPa to be unstable and difficult to parameterize. The \VPa tends to work against the trunk flexion when stepping down, while \VPb assists to the natural response.

To adjust the damping coefficient $c$, we use the duty factor as the primary criteria. Duty factor equals to the stance time over the stride and it decreases with the running speed in human running, as shown with green lines in \cref{fig:duty} \cite{Bishop_2017,Gatesy_1991}. 
We impose the same relation when tuning our gaits for level running, where the duty factor of our level running gaits range between 40-\SI{28}{\percent} for \VPa and 35-\SI{25}{\percent} for \VPb (see \cref{fig:duty}a).
When the terrain has a single step-down, the control scheme attenuates the perturbation and brings the system back to its initial equilibrium state to the same duty factor level.
In case of a downhill terrain, the duty factor decreases proportional to the terrain grade and ranges between 32-\SI{20}{\percent} for \VPb, which is shown in \cref{fig:duty}b. 
A lower duty factor indicates an increase in peak vertical GRF, which can be prevented with additional parameter adjustments such as decreasing the leg stiffness. 

\begin{figure}[!t]
\centering
\begin{annotatedFigure}
	{\includegraphics[width=0.98\linewidth]{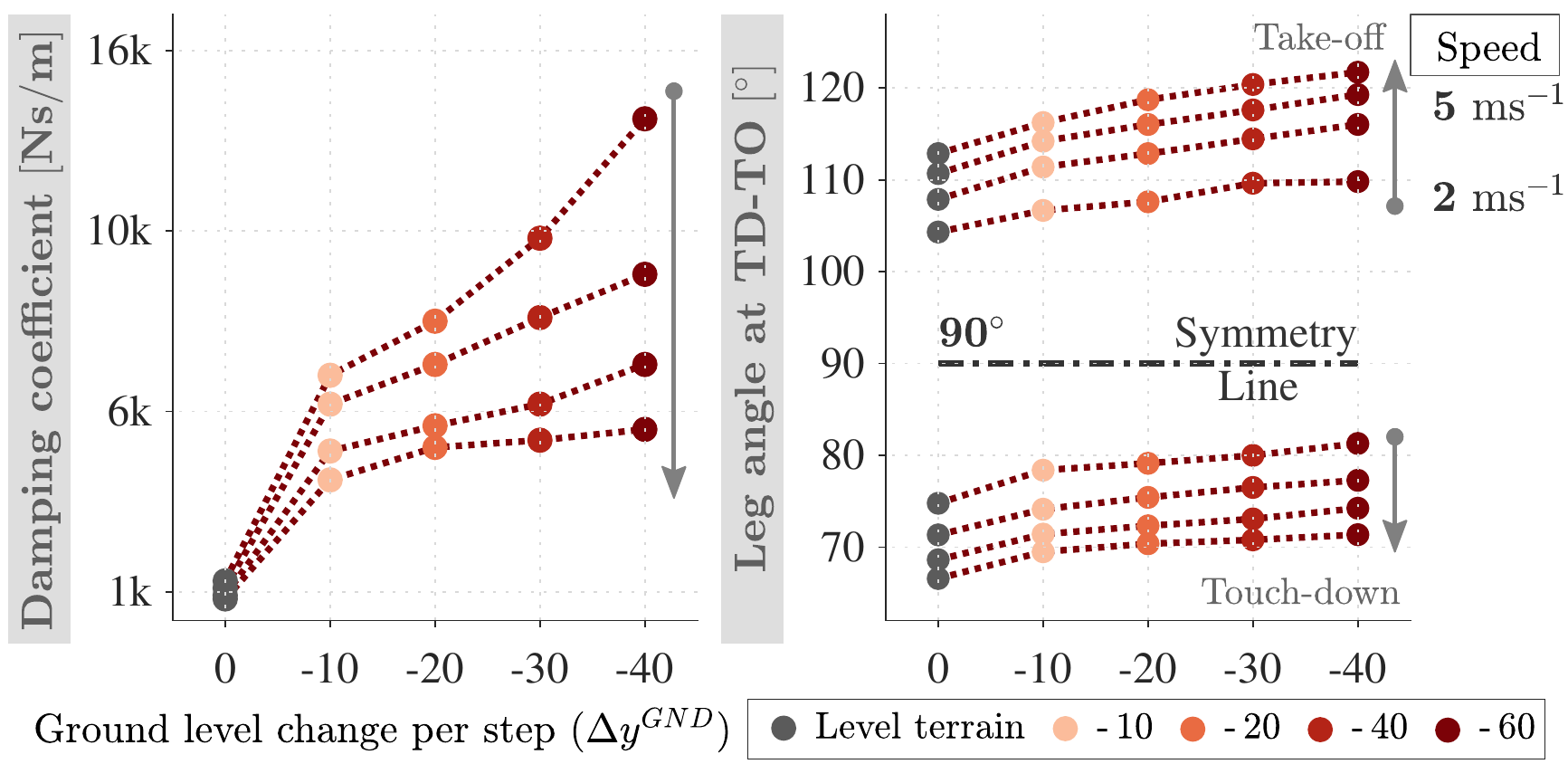}}
	\sublabel{a)}{0.08,0.98}{color_gray}{0.6}   
	 \sublabel{b)}{0.54,0.98}{color_gray}{0.6} 
\end{annotatedFigure}
\caption{ Leg damping coefficient (a) and the leg angle at touch-down/take-off events (b) corresponding to downhill running of speeds 2-5 \si{\meter\per\second} and gradients 0-40 \si{\centi\meter} per step. Damping coefficient and leg angle at touch-down decrease with the speed and increase with the terrain grade. 
%
}\label{fig:Damping}
\vspace{-0.1cm}
\end{figure}
%

%
The second criteria we consider is related to the take-off conditions. If the damping coefficient is too large, the vertical CoM acceleration becomes zero before the leg reaches to its rest length or the vertical GRF reaches to zero. The stance phase is terminated early with GRF suddenly cut-off, and take-off to apex phase of the respective step does not happen. To avoid this unrealistic scenario, we limit the maximum value of the damping coefficient.
Given these considerations we obtain the damping coefficients in \cref{fig:Damping}a, which decrease with speed and increase by a factor of 5-8 with the terrain grade.
The leg angle touch-down exhibits a similar relation to leg damping coefficient, where it decreases with the speed and increases 4-\SI{7}{\percent} with the terrain grade, as shown in \cref{fig:Damping}b.

In the case of the single step-down terrain, the control approach rejects the perturbation in the following level-terrain steps, and returns the system to its initial equilibrium. 
At the downhill terrain, the controller finds a new equilibrium with step-wise disturbance rejection. 
At increasing terrain slope, we observe an increase in the asymmetry of the gait patterns (see \cref{fig:GaitGen}a-\labelcref{fig:GaitGen}b, an asymmetry in the CoM trajectory and GRF).

\section{SIMULATION RESULTS} \label{sec:SimuationResults}
%
In this section, we show how our VP controller responds to the changes in the ground level. We describe the kinetic properties of the gaits and the work distribution between the leg and hip.

\vspace{-0.2cm}
\subsection{Terrain with a Single Step-down}\label{subsec:stepdown}
%
\begin{figure*}[!tbh]
\centering
\begin{annotatedFigure}
	{\includegraphics[width=0.98\linewidth]{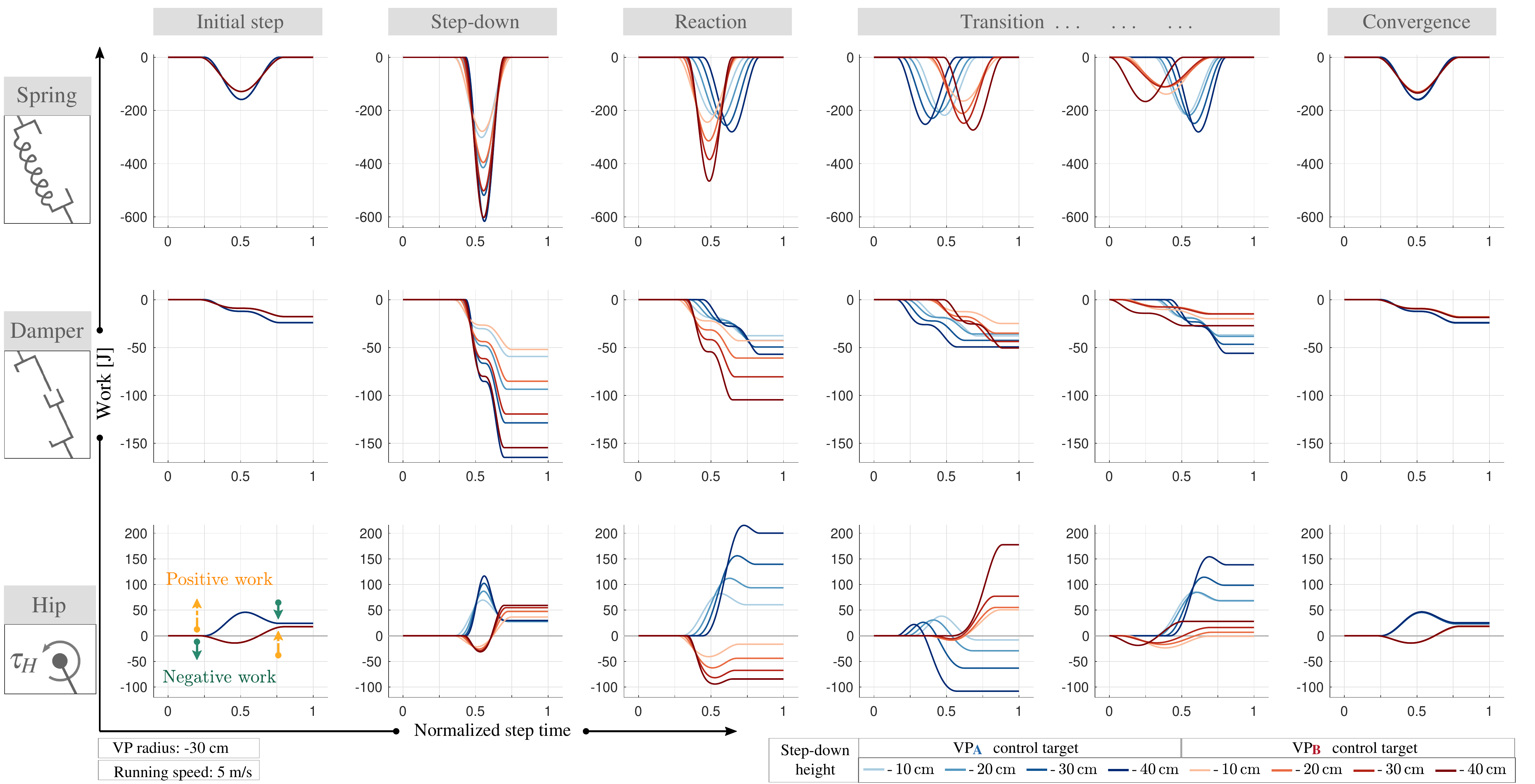}}
	\sublabel{$a_{0}$)}{0.08,0.93}{color_gray}{0.6}    \sublabel{$a_{1}$)}{0.23,0.93}{color_gray}{0.6} 
	\sublabel{$a_{2}$)}{0.385,0.93}{color_gray}{0.6} 	\sublabel{$a_{3}$)}{0.54,0.93}{color_gray}{0.6}
	\sublabel{$a_{4}$)}{0.69,0.93}{color_gray}{0.6} 	\sublabel{$a_{5}$)}{0.845,0.93}{color_gray}{0.6}
	\sublabel{$b_{0}$)}{0.08,0.62}{color_gray}{0.6}    \sublabel{$b_{1}$)}{0.23,0.62}{color_gray}{0.6} 
	\sublabel{$b_{2}$)}{0.385,0.62}{color_gray}{0.6}  \sublabel{$b_{3}$)}{0.54,0.62}{color_gray}{0.6}
	\sublabel{$b_{4}$)}{0.69,0.62}{color_gray}{0.6}  \sublabel{$b_{5}$)}{0.845,0.62}{color_gray}{0.6}
	\sublabel{$c_{0}$)}{0.08,0.34}{color_gray}{0.6}    \sublabel{$c_{1}$)}{0.23,0.34}{color_gray}{0.6} 
	\sublabel{$c_{2}$)}{0.385,0.34}{color_gray}{0.6}  \sublabel{$c_{3}$)}{0.54,0.34}{color_gray}{0.6} 
    \sublabel{$c_{4}$)}{0.69,0.34}{color_gray}{0.6}  \sublabel{$c_{5}$)}{0.845,0.34}{color_gray}{0.6} 
\end{annotatedFigure}
\caption{ The work performed by the leg spring (a), leg damper (b) and the hip actuator (c) during running at \SI{5}{\meter\per\second} for both \VPa and \VPb control targets with \protect\SI{30}{\centi\meter} radius.
The energy of the system increases with the step-down.
The maximum leg compression increases, which leads to an increase in the energy stored/recoiled by the leg spring($a_{1}$) and the energy dissipated by the leg damper ($b_{1}$). 
The net energy generated by the hip increases as well ($c_{1}$). 
In the next step, the VP control starts reacting to the changes in state ($a_{2} \protect\minus c_{2}$). 
In the subsequent steps, the net leg work gradually decreases ($b_{2} \protect\minus b_{4}$), whereas the net hip work alternates its sign to regulate the excess energy ($c_{2} \protect\minus c_{4}$). 
Bot \VPa and \VPb control targets are able to attenuate the excess energy introduced by the step-down and bring the system back to its initial equilibrium conditions within 50 steps.
%
}\label{fig:StepDown_Work_VP}
\end{figure*}

The initial step before the step-down is in equilibrium state for level-terrain, where the leg removes energy from the system and the hip supplies an equal amount of energy, as shown in \cref{fig:StepDown_Work_VP}$a_{0}$-\labelcref{fig:StepDown_Work_VP}$c_{0}$ for \SI{5}{\meter\per\second} speed.
The energy provided by the hip actuator depends on the position of the VP. 
When the control target is \VPa, the hip produces energy at early stance and dissipates energy partially after mid-stance, which results in a net positive work that is required to counterbalance the leg damper (blue lines in \cref{fig:StepDown_Work_VP}c). 
Conversely, the hip actuator with \VPb control target  dissipates energy first and generates a large amount of energy afterwards to compensate for both the prior loss and leg damping (red lines in \cref{fig:StepDown_Work_VP}c).

At step-down perturbation, the total energy of the system increases proportional to the step height, which disrupts the energy balance of the system.
Since the perturbation is one-time-only, the controller has the opportunity to dissipate the perturbation in the following multiple steps, unlike downhill running where the perturbation is continuous and needs to be dissipated within a single step. 
In both cases, the additional energy can not be converted to kinetic energy, since the leg angle controller attempts to maintain a constant running speed and constant trunk angular excursion. 
As a consequence, the excess energy needs to be dissipated through the interplay between in the leg and the hip.
%

During the step-down, the maximum leg compression increases by a factor of 1.3-2 for \VPa and 1.5-2.2 for \VPb, which leads to an increase in the energy stored/recoiled by the spring by a factor of 1.9-3.8 for \VPa and \VPb (see \cref{fig:StepDown_Work_VP}$a_{1}$).
%
Alongside the spring, the energy dissipated by the leg damper increases by 6-11 times higher for \VPa and 7-13 times for \VPb (see \labelcref{fig:StepDown_Work_VP}$b_{1}$).
Both the energy stored/recoiled by spring and the energy dissipated by damper increase with step-down height and running speed.
%
%
The net hip work increases by a factor of 5.1-5.2 for \VPa and 6-7.3 for \VPb. In addition, the peak positive hip work gets 3.6-4.6 times higher for \VPa, whereas the peak negative hip work is 4.6 times larger for \VPb (see \labelcref{fig:StepDown_Work_VP}$c_{1}$).
The VP position update takes place at the end of the step-down, and the controller reacts to the changes in the state in the next steps. 
In the following steps, we observe leg and hip energy fluctuations, where the net damping energy decreases (see \labelcref{fig:StepDown_Work_VP}$b_{2}$-\labelcref{fig:StepDown_Work_VP}$b_{4}$) and net hip energy alternates its sign over the subsequent steps (see \labelcref{fig:StepDown_Work_VP}$c_{2}$-\labelcref{fig:StepDown_Work_VP}$c_{4}$).
In addition, we observe a temporal shift in the stance phase, which alternates over the course of the transition period.

We report gait parameter combinations for both \VPa and \VPb approaches, where the controllers are able to bring the system back to its initial equilibrium conditions within 50 steps (see \cref{fig:StepDown_Work_VP}$a_{5}$-\labelcref{fig:StepDown_Work_VP}$c_{5}$).
%


\begin{figure}[tb!]
\centering
\begin{annotatedFigure}
	{\includegraphics[width=0.98\linewidth]{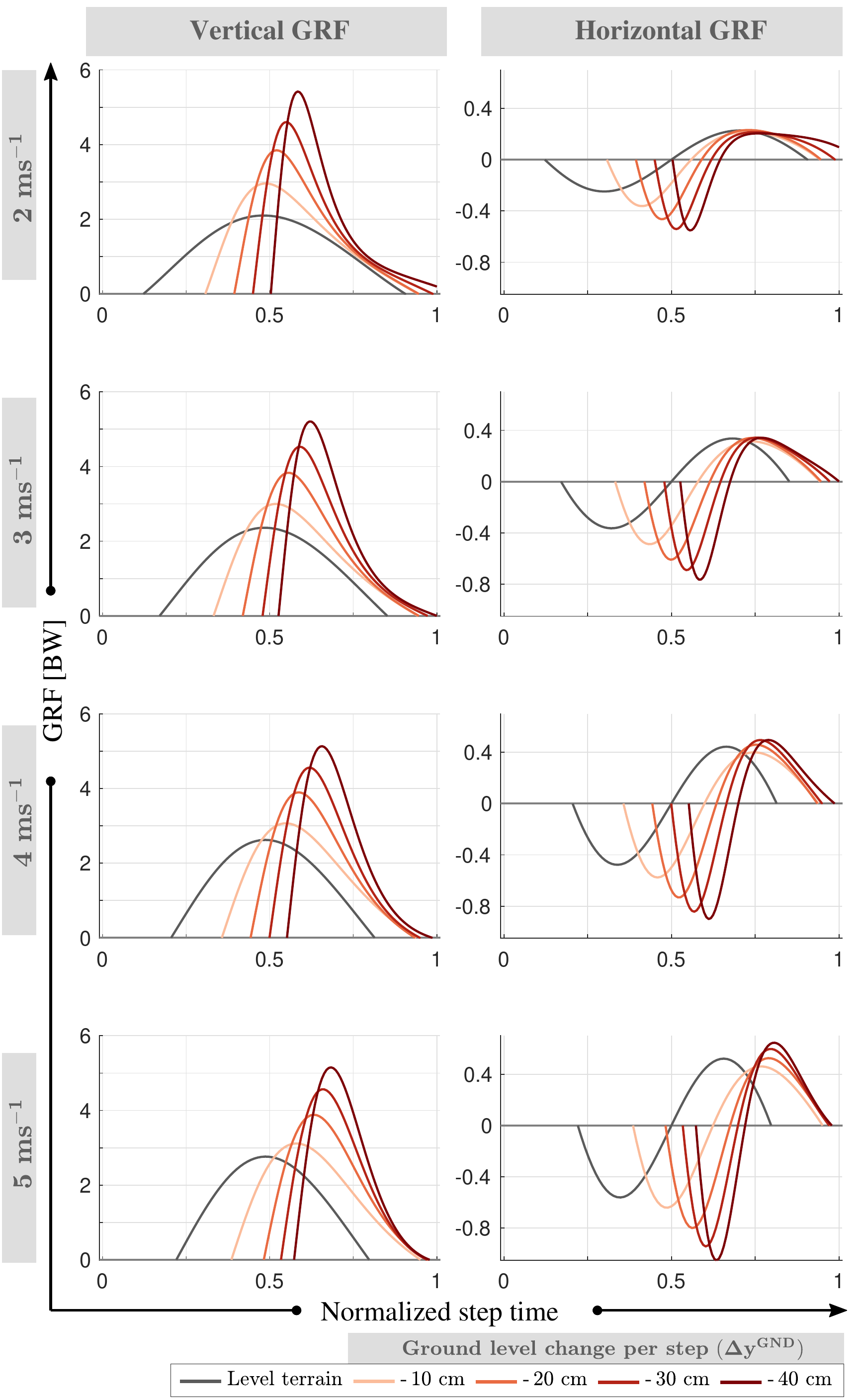}}
	\sublabel{$a_{1}$)}{0.145,0.95}{color_gray}{0.6}    \sublabel{$b_{1}$)}{0.61,0.95}{color_gray}{0.6} 
	\sublabel{$a_{2}$)}{0.145,0.72}{color_gray}{0.6}    \sublabel{$b_{2}$)}{0.61,0.72}{color_gray}{0.6} 
	\sublabel{$a_{3}$)}{0.145,0.49}{color_gray}{0.6}    \sublabel{$b_{3}$)}{0.61,0.49}{color_gray}{0.6} 
	\sublabel{$a_{4}$)}{0.145,0.26}{color_gray}{0.6}    \sublabel{$b_{4}$)}{0.61,0.26}{color_gray}{0.6} 
\end{annotatedFigure}
\caption{ Normalized vertical (a) and horizontal (b) ground reaction forces for downhill running with \VPb control target at speeds of 2-\SI{5}{\meter\per\second}. 
The peak vertical GRF increase with speed and increasing terrain grade. The peak braking forces (min.~horizontal GRF) and peak propulsion forces (max.~horizontal GRF) show a similar behavior. An exception is the peak propulsion forces at $\Delta y^{\mysup{GND}} \myeq$\SI{10}{\centi\meter} ($b_{1}$), which decrease with respect to the level terrain conditions.
In addition, the stance phases shift towards the end of the step and the GRF profiles become more left-skewed with higher terrain grade.
%
}\label{fig:Downill_GRF_VPb}
\end{figure}

\subsection{Downhill Terrain}\label{subsec:downhill}

In downhill running the biggest challenge is to reject the energy introduced by the ground level change within a single step. The controller needs to bring the system to a new equilibrium, where the energy increase due to step-down is dissipated within a single stance phase.

\begin{figure}[tb!]
\centering
\begin{annotatedFigure}
	{\includegraphics[width=0.98\linewidth]{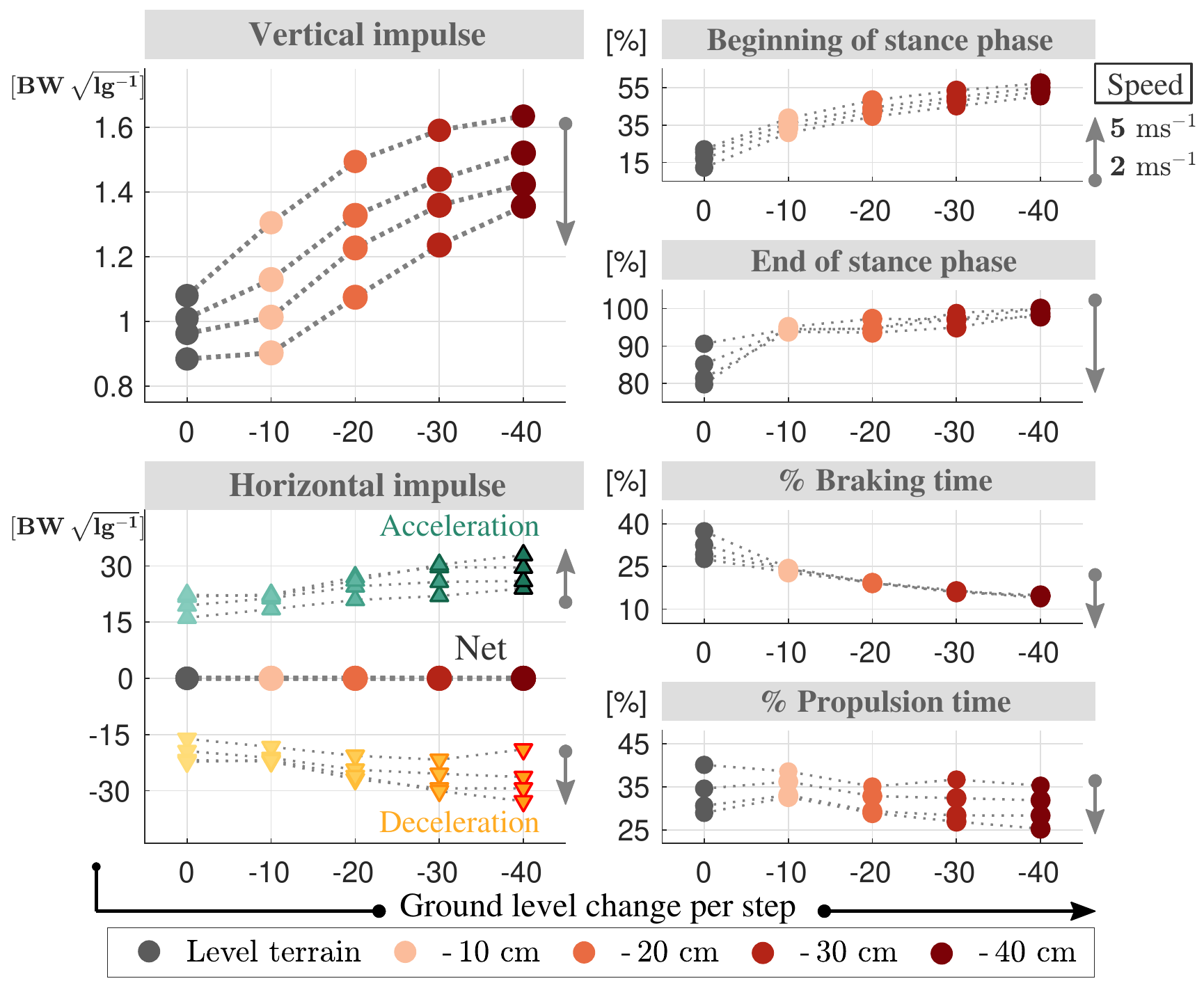}}
	\sublabel{$a$)}{0.15,0.92}{color_gray}{0.6}    \sublabel{$b_{1}$)}{0.58,0.92}{color_gray}{0.6}   \sublabel{$b_{2}$)}{0.58,0.735}{color_gray}{0.6} 
	\sublabel{$c$)}{0.15,0.46}{color_gray}{0.6}    \sublabel{$d_{1}$)}{0.58,0.51}{color_gray}{0.6}   \sublabel{$d_{2}$)}{0.58,0.285}{color_gray}{0.6} 
\end{annotatedFigure}
\caption{Normalized vertical (a) and horizontal (c)  impulses for downhill running with \VPb control target at speeds of 2-\SI{5}{\meter\per\second}. The vertical impulse increases with terrain grade and decreases with speed. Both braking and propulsion impulses increase with the terrain grade and speed, where the sum is equivalent to zero.
The temporal analysis is provided in the right column, where the beginning ($b_{1}$) and end ($b_{2}$) of stance phase over step time is presented in \%, as well as the times spent on braking ($d_{1}$) and propulsion ($d_{3}$)  intervals. The stance time is shifted towards the end of step and  braking /propulsion intervals get shorter, as the terrain grade increases.
%
}\label{fig:Downill_Impulse_VPb}
\end{figure}

To characterize the new equilibrium conditions corresponding to different downhill grades, we evaluate the GRF profiles and impulses. 
%
The peak vertical GRF depends on the running speed and increases from 2 to 2.7 body weights as the speed rises from 2 to \SI{5}{\meter\per\second} in level running (gray lines in \cref{fig:Downill_GRF_VPb}a).
At downhill terrain, the peak vertical GRF increases by a factor of 1.2-2.5  proportional to terrain grade, which reaches up to 5.4 body weights (see \labelcref{fig:Downill_GRF_VPb}$a_{1}$-$a_{4}$). 
The impulses corresponding to the vertical GRF are quantified in \cref{fig:Downill_Impulse_VPb}$a$, which range between 0.9-1.1 for level running and increase from 0.9-1.3 to 1.3-1.6  with the terrain grade.  
In addition, we observe left-skewed vertical GRF profiles, where the asymmetry becomes more pronounced as the terrain grade increases.

The literature in human running provides different answers on how the horizontal GRF responds to downhill terrain conditions. Studies by \cite{Dick_1987,Gottschall_2005,Wells_2018} observe an increase in peak braking forces (min. horizontal GRF) and a decrease in peak propulsion forces (max. horizontal GRF), whereas \cite{Telhan_2010,Yokozawa_2005} report no changes. 
In our level gaits, the peak horizontal GRF magnitude increases with speed and range between 0.22-0.56 body-weights.
At downhill terrain, the peak braking force increases by a factor of 1.1-2.2 with the terrain grade (see \cref{fig:Downill_GRF_VPb}$b_{1}$-$b_{4}$). 
The peak propulsion forces decrease by a factor of 0.96-0.88 when the terrain grad is $\Delta y^{\mysup{GND}} \myeq$\,-\,\SI{10}{cm}, where they increase by a factor of 1.0-1.3 for higher grades.
This dependence on terrain grade could possibly be related to the metabolic minimum observed in \SI{20}{\percent} downhill grade in human running \cite{Minetti_2002,Vernillo_2017}, but no relevant experimental data exists yet.
The peak braking forces being larger than the peak propulsion forces raises the question whether the net horizontal GRF impulse is negative valued to compensate the downhill conditions. 
We see in  \cref{fig:Downill_Impulse_VPb}$c$ that this is not the case. Both the braking and propulsion impulse becomes 1.0-1.5 higher with terrain grade, while the sum remains zero. 
In other words, there is no net horizontal acceleration in our downhill running gaits. 
In addition, we observe a left-skew in the braking and propulsion force patterns, similar to the vertical GRF profiles.

\begin{figure}[!tb]
\centering
\begin{annotatedFigure}
	{\includegraphics[width=0.98\linewidth]{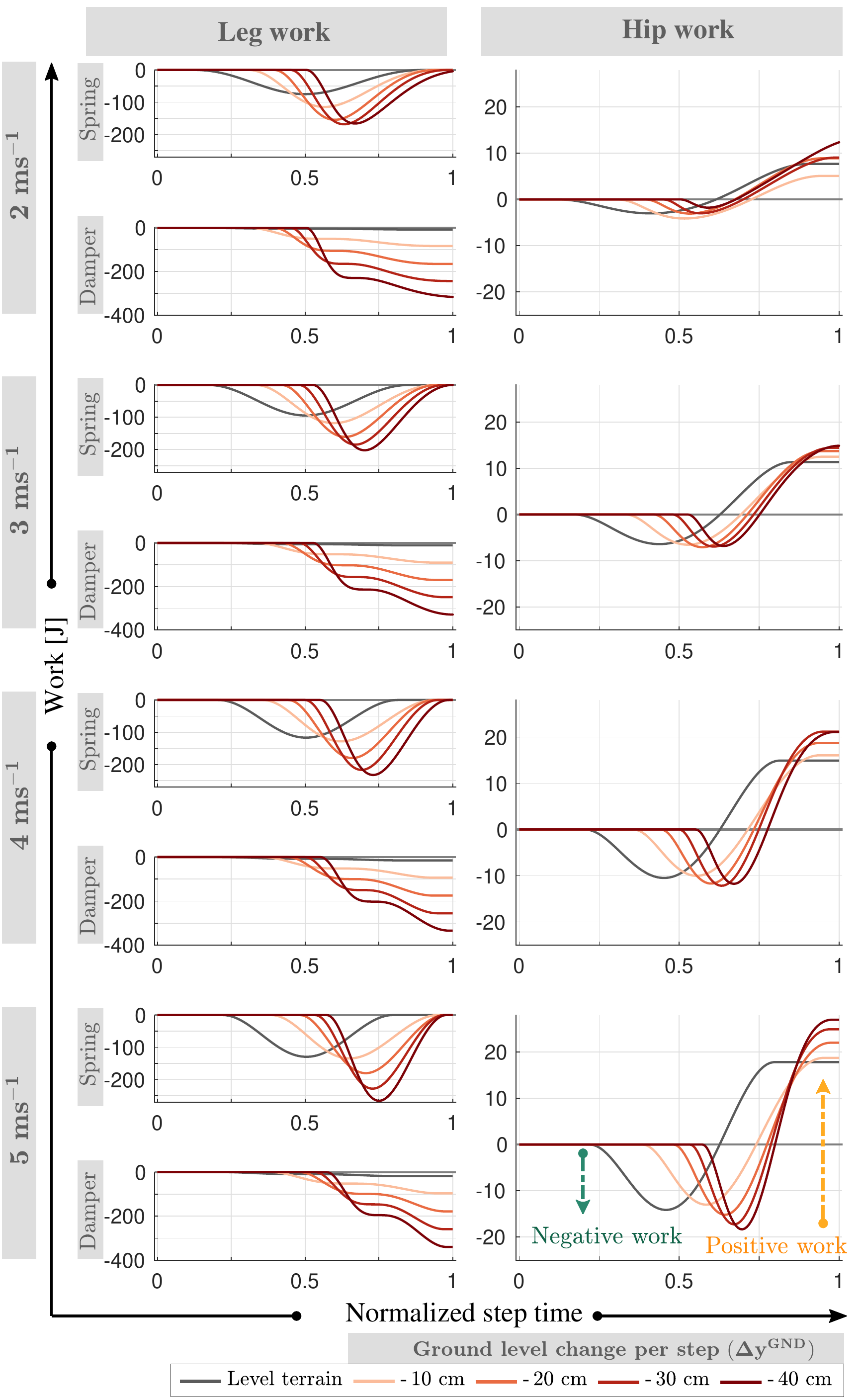}}
\end{annotatedFigure}
\caption{ The work performed by the leg (left) and hip (right) for downhill running with a \VPb control target at speeds of 2-\SI{5}{\meter\per\second}. The stored/recoiled leg spring energy and dissipated damping energy increase with terrain grade and speed. The negative hip work and net energy generated by the hip follows a similar trend.
%
}\label{fig:Downill_Work_dt_VPb}
\end{figure}

To analyze the asymmetric behavior that we observe in the GRF patterns, we analyze the gait’s horizontal and vertical impulses in \cref{fig:Downill_Impulse_VPb}$b$-\labelcref{fig:Downill_Impulse_VPb}$d$. 
In level running, the stance phase begins at 12-\SI{11}{\percent} of the step and ends at 90-\SI{79}{\percent}, where the braking/propulsion intervals comprise 37-\SI{27}{\percent} and 40-\SI{29}{\percent} of the step time respectively. As the downhill terrain grade increases, the stance phase shifts towards the end of step, while the braking/propulsion intervals decrease. For the grade $\Delta y^{\mysup{GND}} \myeq$\,-\,\SI{40}{\centi\meter}, the stance phase starts at 55-\SI{57}{\percent} and continues until the end of the step time. In this case, the phase between leg take-off and the apex diminishes.

\begin{figure}[!t]
\centering
\begin{annotatedFigure}
	{\includegraphics[width=0.98\linewidth]{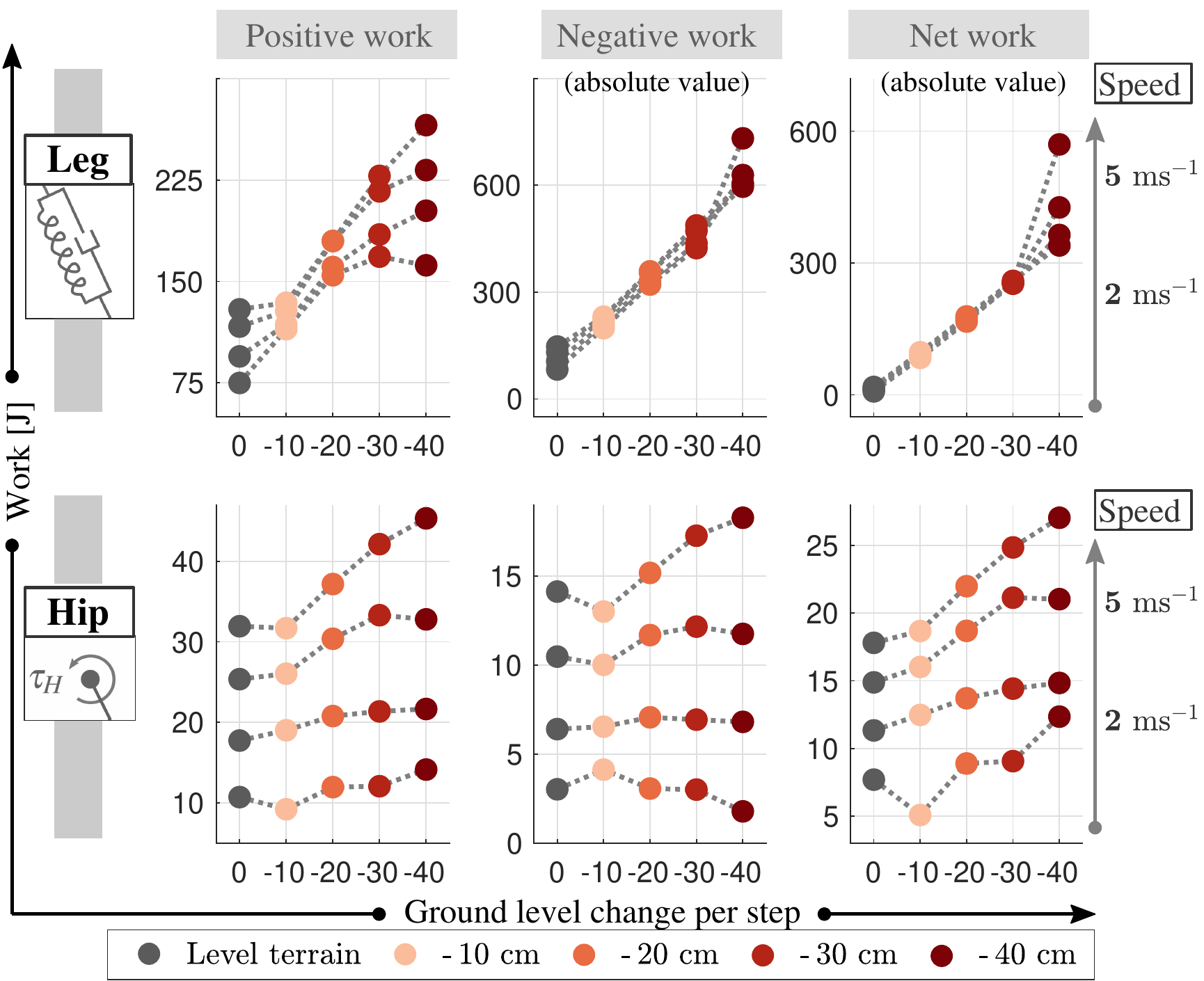}}
	\sublabel{$a_{1}$)}{0.16,0.93}{color_gray}{0.6}    \sublabel{$a_{2}$)}{0.42,0.93}{color_gray}{0.6}  \sublabel{$a_{3}$)}{0.68,0.93}{color_gray}{0.6} 
		\sublabel{$b_{1}$)}{0.16,0.5}{color_gray}{0.6}    \sublabel{$b_{2}$)}{0.42,0.5}{color_gray}{0.6}  \sublabel{$b_{3}$)}{0.68,0.5}{color_gray}{0.6} 
\end{annotatedFigure}
\caption{ The positive ($a_{1},b_{1}$), negative ($a_{2},b_{2}$) and net ($a_{3},b_{3}$) work performed by the leg and hip for downhill running with a \VPb control target at speeds of 2-\SI{5}{\meter\per\second}. The increase of energy caused the downhill terrain is compensated by the increase in the energy dissipated in leg and the energy generated by hip increases.
%
}\label{fig:Downill_Work_Net_VPb}
\end{figure}

\begin{figure}[!t]
\centering
\begin{annotatedFigure}
	{\includegraphics[width=0.98\linewidth]{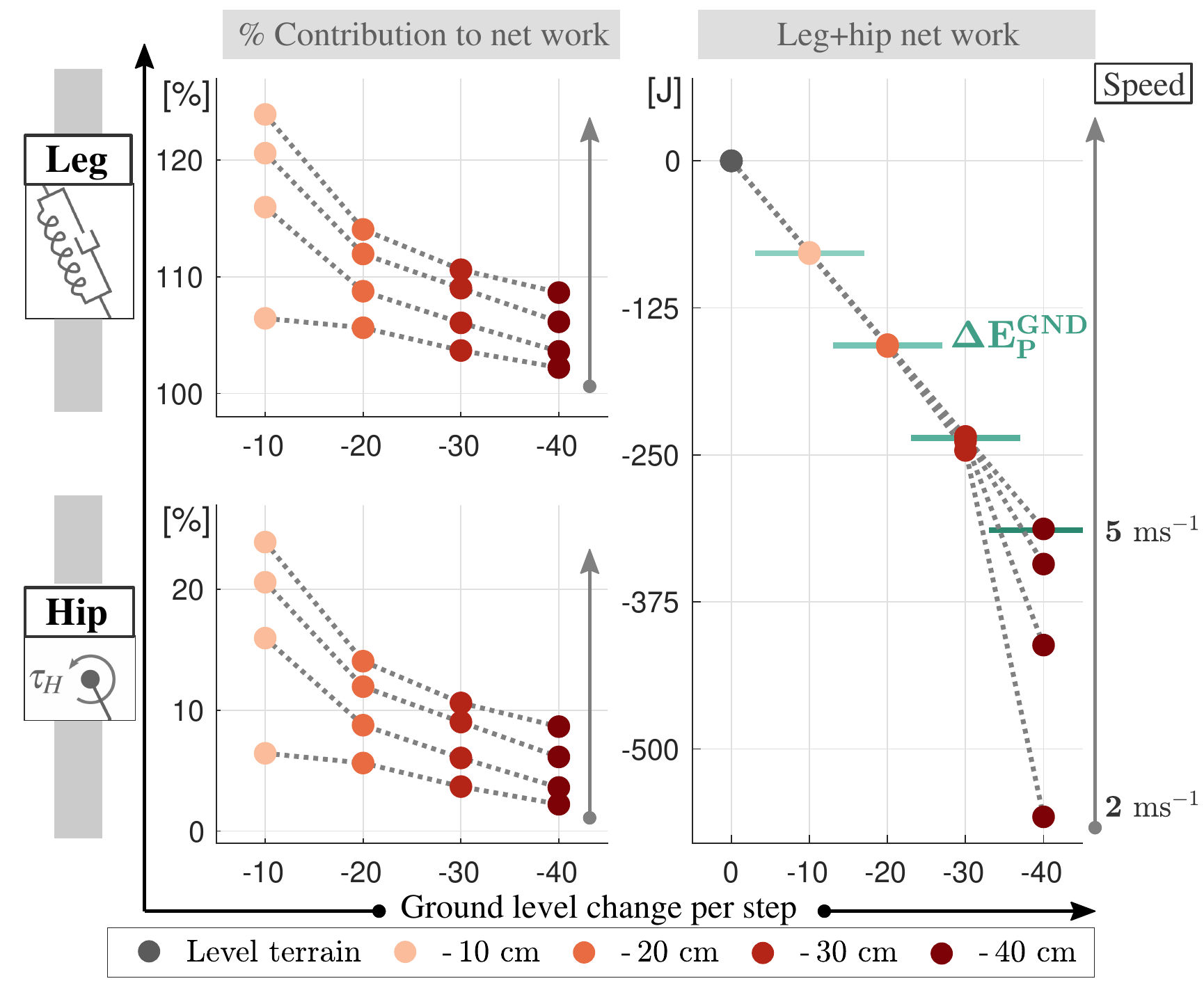}}
	\sublabel{a)}{0.16,0.95}{color_gray}{0.6}   
	\sublabel{b)}{0.16,0.51}{color_gray}{0.6}   
	\sublabel{c)}{0.55,0.95}{color_gray}{0.6}   
\end{annotatedFigure}
\caption{Work contribution of the leg (a) and hip (b) to the net work (in percent) and the numerical values of the net work corresponding to varying ground level changes (c). The leg removes energy from the system, whereas the hip injects energy. The amount of potential energy added to the system at the downhill conditions ($\Delta E_{\mysub{P}}^{\mysup{GND}}$) is marked with solid green lines in (c). We see that the increase from additional potential energy is fully compensated by the leg-hip actuators for shallow down-slopes, where $| \Delta y^{\mysup{GND}}| \leq$ \SI{30}{\centi\meter}.
%
}\label{fig:Downill_Work_Contribution_VPb}
\end{figure}

Next, we investigate how the VP control compensates for the additional energy caused by the ground level changes. Unlike \cite{Kenwright_2011} who suggests to offset the VP position horizontally, we found it sufficient to increase the damping coefficient to accommodate downhill grades.
The time progression of the work performed by the leg and hip is shown in \cref{fig:Downill_Work_dt_VPb} and corresponding numerical values for the positive/negative/net works are given in \cref{fig:Downill_Work_Net_VPb}.
As the terrain grade increases, the positive leg work increases by a factor of 1.5-2.2 (see \labelcref{fig:Downill_Work_Net_VPb}$a_{1}$), negative leg work by a factor of 1.6-6 (see \labelcref{fig:Downill_Work_Net_VPb}$a_{2}$), and the net leg work by a factor of 5.5-37 (see \labelcref{fig:Downill_Work_Net_VPb}$a_{3}$).
On the other hand, the positive hip work gets 1.02-2.4 times higher (see \labelcref{fig:Downill_Work_Net_VPb}$b_{1}$), negative hip work gets 1.01-1.3 times higher with the exception of $\Delta y^{\mysup{GND}}\myeq$\,-\,\SI{10}{\centi\meter} (see \labelcref{fig:Downill_Work_Net_VPb}$b_{2}$), and net hip work gets 1.1-1.5 times higher with the terrain grade (see \labelcref{fig:Downill_Work_Net_VPb}$b_{3}$).

In \cref{fig:Downill_Work_Contribution_VPb}, we look at the distribution of the net work provided by the leg and hip actuators. The sum of the energy dissipated by the leg (see \cref{fig:Downill_Work_Contribution_VPb}a) and produced by the hip (see \cref{fig:Downill_Work_Contribution_VPb}b) amounts to \SI{100}{\percent}. When the terrain grade increases, the percentage contributions of the leg and hip decrease. The combined work of the leg and hip dissipates the energy introduced by the terrain's potential energy difference (see $\Delta E_{\mysub{P}}^{\mysup{GND}}$ in \cref{fig:Downill_Work_Contribution_VPb}c). The relation holds for terrain grades up to $|\Delta y^{\mysup{GND}}| \leq $ \SI{30}{\centi\meter}, whereas for higher grades dissipated energy is larger than the terrain's potential energy change.


\section{CONCLUSION}\label{sec:Conclusion}

In this work, we investigated the virtual point control mechanism in its ability to cope with single step-down perturbations and downhill terrains, using a simple spring inverted pendulum model with trunk.
We showed that placing the virtual point either above or below the center of mass allows to dissipate the perturbations caused by a single step-down in terrain up to a step height of \SI{40}{\centi\meter} at speeds of 2-\SI{5}{\meter\per\second}.
In addition, we found that increasing the leg damping and placing the virtual point below the center of mass is sufficient to compensate for the energetic and dynamic changes introduced by downhill running. 
No further virtual point manipulation was necessary. 
Our results provide an easy recipe to parameterize humanoid robot controllers, to adjust for varying terrain conditions.
%
%


 \bibliographystyle{IEEEtran}
\bibliography{root.bib}



\end{document}